\documentclass[times,twocolumn,final,authoryear]{elsarticle}

\usepackage{ycviu}
\usepackage{framed,multirow}

\usepackage{amssymb}
\usepackage{latexsym}

\usepackage{url}
\usepackage{xcolor}
\definecolor{newcolor}{rgb}{.8,.349,.1}
\definecolor{mygray}{rgb}{0.85, 0.85, 0.85}

\usepackage{amsmath,amsfonts}
\usepackage{algorithmic}
\usepackage{algorithm}
\usepackage{array}
\usepackage{textcomp}
\usepackage{verbatim}
\usepackage{graphicx}

\usepackage{subfig}
\usepackage{makecell}
\usepackage{booktabs}
\usepackage{multirow}

\newcommand{\ie}{\textit{i}.\textit{e}.}
\newcommand{\eg}{\textit{e}.\textit{g}.}
\newcommand{\etc}{\textit{etc}}

\journal{Computer Vision and Image Understanding}

\begin{document}

\clearpage

\ifpreprint
  \setcounter{page}{1}
\else
  \setcounter{page}{1}
\fi

\begin{frontmatter}

\title{Hierarchical compositional representations for few-shot action recognition}

\author[1,2,3]{Changzhen Li} 
\author[1,2]{Jie Zhang}

\author[4]{Shuzhe Wu}
\author[4]{Xin Jin}
\author[1,2,3]{Shiguang Shan}

\address[1]{Key Lab of Intelligent Information Processing of Chinese Academy of Sciences (CAS), Institute of Computing Technology, CAS, Beijing 100190, China}
\address[2]{University of Chinese Academy of Sciences, Beijing 100049, China}
\address[3]{Hangzhou Institute for Advanced Study, UCAS, school of Intelligent Science and Technology}
\address[4]{Beijing Huawei Cloud Computing Technologies Co., Ltd, No. 3 Xinxi Road, Haidian District, Beijing 100095, China}

\received{1 May 2013}
\finalform{10 May 2013}
\accepted{13 May 2013}
\availableonline{15 May 2013}
\communicated{S. Sarkar}

\begin{abstract}
Recently action recognition has received more and more attention for its comprehensive and practical applications in intelligent surveillance and human-computer interaction. However, few-shot action recognition has not been well explored and remains challenging because of data scarcity.
In this paper, we propose a novel hierarchical compositional representations (HCR) learning approach for few-shot action recognition.
Specifically, we divide a complicated action into several sub-actions by carefully designed hierarchical clustering and further decompose the sub-actions into more fine-grained spatially attentional sub-actions (SAS-actions). Although there exist large differences between base classes and novel classes, they can share similar patterns in sub-actions or SAS-actions. Furthermore, we adopt the Earth Mover’s Distance in the transportation problem to measure the similarity between video samples in terms of sub-action representations. It computes the optimal matching flows between sub-actions as distance metric, which is favorable for comparing fine-grained patterns. Extensive experiments show our method achieves the state-of-the-art results on HMDB51, UCF101 and Kinetics datasets.
\end{abstract}

\begin{keyword}
\MSC 41A05\sep 41A10\sep 65D05\sep 65D17
\KWD Keyword1\sep Keyword2\sep Keyword3

%% MSC codes here, in the form: \MSC code \sep code
%% or \MSC[2008] code \sep code (2000 is the default)
\end{keyword}

\end{frontmatter}

%\linenumbers

%% main text
\section{Introduction}
Action recognition has received considerable attention with the explosive increase of videos in daily life. Recent action recognition methods benefit from the significant progress of Convolutional Neural Networks (CNNs) \citep{krizhevsky2012imagenet,he2016deep,simonyan2014two,tran2015learning}, which is much dependent on massive labeled data. 
However, there exist two problems of collecting massive data for action recognition:
1) collecting and annotating videos is extremely expensive as it takes much time to watch all the frames. 
2) it is almost impossible to collect all potential action classes, since human actions are abundant and dynamic, \eg, the ``balance beam'' consists of up to 193 fine-grained classes based on its jump style \citep{shao2020finegym}. 
So few-shot action recognition (FSAR), as learning to recognize actions of novel classes with only few labeled samples, is quite valuable for real-world applications. 

\vspace{45pt}

\begin{figure}[!t]
    \centering
    \includegraphics[width=0.45\textwidth]{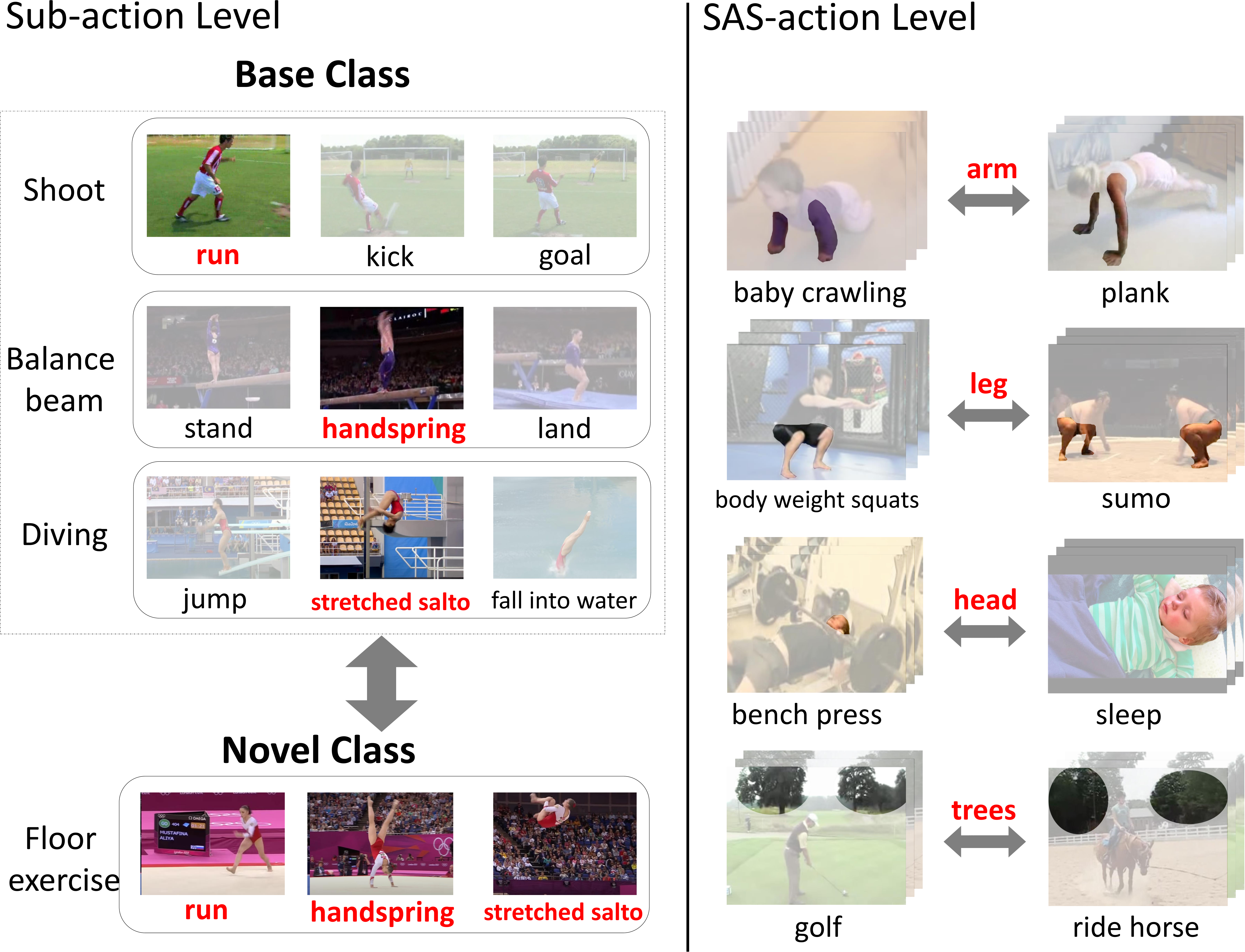}
    \caption{Although there exist differences between base classes and novel classes, they can share basic patterns in common, \eg, sub-actions and SAS-actions.}
    \label{fig:1_1}
    \vspace{20pt}
\end{figure}

Few-shot learning (FSL) aims to address the learning problem with limited training samples. Representative metric-based methods in FSL \citep{vinyals2016matching,snell2017prototypical,sung2018learning,zhang2020deepemd} have achieved promising results that map samples to an appropriate embedding space and measure the similarity between support and query samples.
Although few-shot learning has achieved great success in image recognition, few-shot action recognition has not been well explored and remains challenging not just because of data scarcity but also the difficulty of temporal modeling. 
Typical few-shot action recognition methods \citep{zhang2020few,zhu2018compound} extract global spatio-temporal representations by temporal pooling operations to measure the similarity between video samples. 
However, the process of computing global representations of the videos may destroy the temporal information and ignore local information, \eg, sub-actions, which probably provides the discriminative information and transferable knowledge \citep{zhang2020deepemd}. 
Recently, some methods \citep{cao2020few} resort to modeling and measuring local representations in term of the clips. 
These methods first divide the video sequence into clips with equal durations following TSN \citep{wang2016temporal}, and further align two local representations by timing sequence explicitly.
However, splitting clips may cut off similar semantics between clips, and meanwhile simple aligning operation is unable to deal with \textbf{time-independent} actions which have no sequential consistency, \eg, for the action ``eat'', Person A has eggs first and then drinks milk, while Person B drinks milk first and then has eggs. (See more details on time-independent actions in Appendix A.). 
Therefore, directly aligning clip representations between video sequences by temporal restriction is not the optimal solution.

Cognitive science indicates that humans' visual system decomposes a complicated thing into simple components and recognizes by components \citep{hoffman1984parts}. 
Following this thought of simplifying things, we divide a complicated action into several simple sub-actions along the temporal dimension and recognize the action by sub-actions. 
However, in the few-shot learning settings, novel classes severely differ from base classes, and even sub-actions between novel and base classes have no similar patterns for transferring.
Therefore, we further decompose the sub-actions into more fine-grained Spatially Attentional Sub-actions (SAS-actions) along the spatial dimension inspired by early works modeling articulated bodies parts \citep{felzenszwalb2005pictorial}.
These SAS-actions include body parts (\ie, arm and leg) and contexts (\ie, trees) as shown in Fig. \ref{fig:1_1}.
Although there exist large gaps between base actions and novel actions, they can share basic SAS-actions in common. \eg, almost all of the actions in HMDB51 \citep{kuehne2011hmdb} contain the SAS-action of "arm move". 
Therefore, we can generalize fine-grained patterns from abundant base classes and transfer them to learn novel classes, and these fine-grained patterns can help provide discriminative and transferable information across categories for classification.

In this paper, we propose a novel hierarchical compositional representations (HCR) learning approach for few-shot action recognition. As shown in Fig. \ref{fig:3_1}, we first divide a complicated action into several sub-actions by clustering, and then further decompose the sub-actions into more fine-grained SAS-actions by carefully designed Parts Attention Module (PAM), which forms hierarchical compositional representations (see detailed visualizations in Appendix C.). 
These fine-grained SAS-actions consist of explicit SAS-actions and implicit SAS-actions. The former corresponds to pre-defined body parts, which are learned by parts prior constraint. The latter corresponds to other action-relevant cues like context.
Different from the existing works \citep{cao2020few}, we modify the conventional hierarchical clustering to partition a video into sub-actions of varying temporal lengths rather than equally splitting a video sequence into clips. Therefore, similar video frames are gathered and continuous semantics are reserved together within sub-actions.
Furthermore, considering that directly aligning local representations along temporal dimension cannot well handle time-independent action samples, we adopt Earth Mover’s Distance (EMD) as distance function to match sub-action representations to better compare fine-grained patterns.
Within clips, the timing sequence is well preserved in clustered sub-actions, and between clips, the timing sequence is ignored by the process of optimizing the EMD distance, which is much suitable for few-shot action recognition. (See more details in Appendix B.)
Finally, after obtaining matching similarity from EMD, we adopt the softmax function to compute the probability over various actions.

To verify the effectiveness of our method, we conduct comprehensive experiments on the popular benchmarks, \ie, HMDB51, UCF101 and Kinetics. In all, our contributions are four-fold: 
\begin{itemize}
    \item We propose the hierarchical compositional representations in terms of fine-grained sub-action and SAS-action components, which encourages to learn more common patterns between novel and base classes. These common fine-grained patterns provide discriminative and transferable information for recognizing novel class examples with only few labeled samples.
    \item We carefully design Parts Attention Module (PAM) to pay attention to various regions of interest, and especially, explicit SAS-actions contain pre-defined human body parts and implicit SAS-actions contain other action-relevant cues like context.
    \item To better compare fine-grained patterns, we adopt Earth Mover's Distance as distance metric for few-shot action recognition to handle time-independent action, which can well match these fine-grained and discriminative sub-action representations.
    \item Extensive experiments show that our method achieves state-of-the-art results on HMDB51, UCF101 and Kinetics datasets.
\end{itemize}

\clearpage
\section{Related Work}
\vspace{-8pt}

\subsection{Action recognition}
\vspace{-4pt}
Traditional approaches, \eg, iDT \citep{wang2013action}, employ hand-crafted spatio-temporal representations for action recognition. After the breakthrough of Convolutional Neural Networks, there exist four types of methods to tackle this task. The first method, Two-stream Networks \citep{simonyan2014two}, processes RGB input and optical flow in different branches separately to learn the spatio-temporal information. Inspirited by this, deeper network architectures \citep{wang2015towards}, two-stream fusion \citep{feichtenhofer2016convolutional,wang2017spatiotemporal}, recurrent neural networks \citep{yue2015beyond,li2018videolstm}, segment-based methods \citep{wang2016temporal}, and multi-stream \citep{choutas2018potion,wu2016harnessing} further explore the two-stream schemes. However, the optical flow is computationally expensive and needs to be computed ahead of time. The second trend is to employ 3D ConvNets \citep{tran2015learning} to model spatio-temporal representations, which shows favorable capabilities for action recognition. Following this clue, deeper backbones \citep{diba2017temporal,tran2017convnet}, 3D convolution factorization \citep{tran2018closer,qiu2017learning}, long-range temporal modeling \citep{zhang2020v4d}, and effective spatio-temporal relations \citep{wang2018non,wang2018videos} are further proposed to recognize actions. 
The third type of method \citep{arnab2021vivit,bertasius2021space,fan2021multiscale} presents pure-transformer based models for action recognition, for its success in image recognition.
Finally, the fourth method attempt to specialize in computational efficiency, such as TSM \citep{lin2019tsm}, X3D \citep{feichtenhofer2020x3d}, TVN \citep{piergiovanni2022tiny}, \etc. Although these methods all achieve promising results, most of them rely on massive training data.
\vspace{-12pt}

\subsection{Few-shot learning}
\vspace{-4pt}
Existing few-shot learning works can be divided into three main approaches: metric based approaches \citep{vinyals2016matching,snell2017prototypical,sung2018learning,tokmakov2019learning}, optimization based approaches \citep{finn2017model,ravi2017optimization}, and data augmentation based approaches \citep{hariharan2017low}. 
Metric based approaches aim to represent samples in an appropriate embedding space and then measure the similarity or distance between samples using a distance/metric function. 
MatchingNet \citep{vinyals2016matching}, ProtoNet \citep{snell2017prototypical}, RelationNet \citep{sung2018learning} employ cosine similarity, Euclidean distance, and CNN-based relation module to estimate the similarity score, respectively. 
And \citep{tokmakov2019learning} uses category-level attribute annotations and measure the similarity between
image examples in compositional representations space.
Optimization based approaches are designed to provide good initialized parameters \citep{finn2017model} or learn an optimizer to output search steps \citep{ravi2017optimization} so that the classifiers can achieve rapid adaption with a small number of examples of novel classes. Data augmentation based approaches \citep{hariharan2017low} generally augment novel class samples by hand-crafted rules or training a generator from the base class's data to handle data scarcity. Moreover, data augmentation based approaches are typically combined with other few-shot learning approaches together in most cases.
\vspace{-12pt}

\begin{figure*}[!ht]
    \centering
    \includegraphics[width=\textwidth]{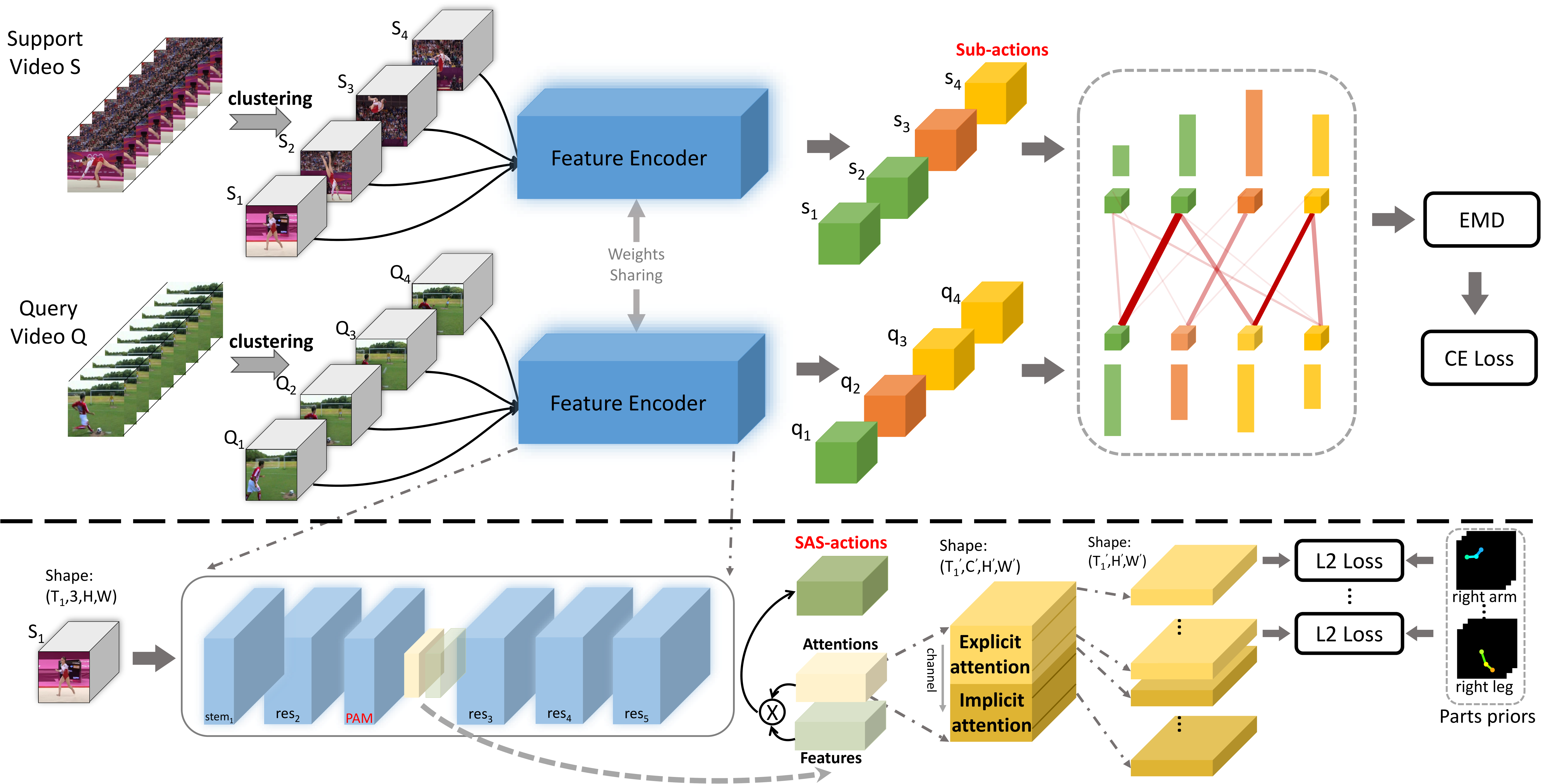}
    \caption{The pipeline. The whole video is first clustered into flexible sub-actions. Each sub-action extracts the corresponding spatio-temporal representations by Feature Encoder. In this process, we regard each channel's output of the Parts Attention Module (PAM) as a SAS-action, and these SAS-actions are further divided into explicit SAS-actions and implicit SAS-actions. The former pays attention to pre-defined human body parts by parts prior constraint, while the latter pays attention to other action-relevant cues like context. Finally, the EMD distance is adopted to measure the distance of sub-action representation sequences between support and query videos.}
    \label{fig:3_1}
\end{figure*}

\subsection{Compositional representation learning} 
\vspace{-1pt}
Compositional representation has been comprehensively studied in cognitive science early \citep{hoffman1984parts}. Traditional approaches in computer vision have inherently embraced compositional methods \citep{marr2010vision} or hierarchical methods \citep{fidler2007towards}. Recently, \cite{tokmakov2019learning} exploited category-level attribute annotations to learn compositional representations in the image domain through deep learning. 
However, the exploration of compositional representation in the video domain has yet to be well explored \citep{lan2015action,malgireddy2013language,ji2020action}. Previously, Mid-level Action Element (MAE) \citep{lan2015action} proposes a hierarchical representation by autonomously parsing videos into mid-level action elements (MAEs) at multiple scales and capturing rich spatial and temporal information through an orchestrated multistage process. Our method further explores the connection between compositional representation learning and deep learning within the video understanding.

\subsection{Few-shot action recognition}
\vspace{-1pt}
Similar to general few-shot learning works, there also exist three main methods for few-shot action recognition, and our approach is closer to metric based approaches. 
In metric based approaches\citep{bishay2019tarn,tan2019learning,cao2020few,perrett2021temporal,bo2020few,lu2021few,li2022ta2n}, TARN \citep{bishay2019tarn} utilizes a segment-by-segment attention mechanism for temporal alignment and learns the distance measure based on the aligned representations. Similarly, FAN \citep{tan2019learning} aligns the representations between video examples and learns a similarity metric by CNNs. 
To better employ the temporal information, OTAM \citep{cao2020few} aligns and measures the distance between video examples using the Dynamic Time Warping (DTW) algorithm. 
Furthermore, TRX \citep{perrett2021temporal} compares the query to sub-sequences of all support set to construct query-specific class prototypes by CrossTransformer attention mechanism.
HCL \citep{zheng2022few} proposes a hierarchical matching approach with a zoom-in matching module that enables comprehensive similarity measurement at global, temporal, and spatial levels, and a mixed-supervised hierarchical contrastive learning (HCL) approach for discriminative temporal and spatial associations. 
MTFAN \citep{wu2022motion} leverages task-specific motion modulation and multi-level temporal fragment alignment to address the challenges posed by the temporal dimension in videos. 
In optimization based approaches \citep{zhu2018compound}, CMN \citep{zhu2018compound} introduce a multi-saliency embedding algorithm to encode video representations and use a compound memory network for classification. In data augmentation based approaches \citep{kumar2019protogan,zhang2020few}, ProtoGAN \citep{kumar2019protogan} synthesizes additional examples for novel categories by conditional generative adversarial networks. ARN \citep{zhang2020few} employs the Attention by Alignment mechanism to localize actions and augments the training data by spatial and temporal self-supervision auxiliary tasks. 
L2A \citep{gowda2022learn2augment} proposes a learned data augmentation approach for video action recognition that uses video compositing to generate diverse and realistic samples while reducing the search space.
Among these, \citep{zhang2020few,bo2020few,zhu2018compound} extract global video-level representations and therefore may destroy the temporal information and ignore local discriminative information.
\citep{bishay2019tarn,tan2019learning, cao2020few} directly align local clip representations by imposing severe temporal restriction and therefore cannot well handle time-independent cases. 
Different from these methods, we design a plain video frames clustering algorithm to generate flexible sub-actions rather than clips with equal durations to alleviate the problem that the sub-action may span across multiple clips.
Besides, we employ the Earth Mover’s Distance (EMD) in the transportation problem to match discriminative local representations rather than to align local representations. The EMD computes the optimal matching flows between two sequences, which can be interpreted as the minimum cost to reconstruct the representations from one to the other \textbf{without} considering the sequential relationship between sub-actions. Meanwhile, the timing sequence within sub-action representations is still preserved.
In short, we keep the timing sequence within sub-actions and discard that between sub-actions.
Therefore, our method is a good trade-off between destroying the temporal information \citep{zhang2020few} and strictly aligning by temporal restriction \citep{cao2020few}, which is much suitable for few-shot action recognition. See more details in Appendix B. 
Overall, our method divides a complicated action into sub-actions, and then further decomposes the sub-actions into more fine-grained SAS-actions. Thus, more similar patterns can be explored between novel and base classes, which is beneficial for transferring knowledge from base to novel actions.
% alleviate the problem of breaking semantic information.

\section{Method}
In this section, we first provide a pipeline of our proposed method named hierarchical compositional
representations (HCR) for few-shot action recognition. Then we give comprehensive introductions of the hierarchical compositional representations and EMD distance metric in our HCR. 
Next, we discuss the difference from existing works. Finally, we describe the implementation details for learning the HCR.

\subsection{Pipeline}
Figure \ref{fig:3_1} shows the pipeline of our hierarchical compositional representations (HCR). First, the raw video $S \in R^{T \times 3 \times H \times W}$ is clustered into $K$ sub-actions $\left\{ {{S_1},{S_2},...,{S_K}} \right\}$ with the varying lengths, where $H$ and $W$ denote the resolution size and T is the total length. 
For each sub-action $S_i \in R^{T_i \times 3 \times H \times W}$, the Feature Encoder extracts the corresponding spatio-temporal representations $s_i$. 
In this process, we regard each channel output of the Parts Attention Module (PAM) as a SAS-action, and these SAS-actions are further divided into explicit SAS-actions and implicit SAS-actions.
Explicit SAS-actions pay attention to pre-defined body parts by parts prior constraint, while implicit SAS-actions pay attention to other action-relevant cues like context.
After a series of convolution and non-linear operation combinations on SAS-action representations, we obtain the sub-action spatio-temporal representations $s_i$. 
Finally, we adopt the EMD distance function to measure the similarity of sub-action representation sequences between support set and query set videos. The final similarity scores are fed to a softmax layer so that they can be mapped to the probability distribution over the sample classes. And the formula is as follows:

\begin{equation}
    \small
    \begin{array}{l}
    \left\{ {{S_1},{S_2},...,{S_K}} \right\}=h\left(S,K\right),  \\
    \{e_1, e_2, \ldots, e_M\}=f\left( {{S_1},{S_2},...,{S_K}} ; \theta_f\right),  \\
    \{s_1, s_2, \ldots s_K\}=g\left(e_1, \ldots, e_M; \theta_g\right),  \\
    \Phi=\left[s_{1}, \ldots s_{K}\right],  \\
    r_{sq} = d\left(\Phi_s,\Phi_q\right).
    \end{array}
\end{equation}
Here, $h$ denotes a frame-based clustering algorithm to generate $K$ sub-actions $\left\{ {{S_1},{S_2},...,{S_K}} \right\}$ with various lengths. $f$ is the function to learn SAS-action representations $\{e_1, e_2, \ldots, e_M\}$ for these sub-actions. $\theta_f$ is the parameters of function $f$ and $M$ denotes the SAS-action number. $g$ is the function to learn sub-action representations $\{s_1, s_2, \ldots, s_K\}$ by merging multiple SAS-actions. $\theta_g$ is the parameters of function $g$ and $K$ denotes the sub-action number. $\Phi$ represents the whole video representation, which consists of a sequence of sub-action representations $\{s_1, s_2, \ldots, s_K\}$. $\Phi_s$ and $\Phi_q$ denote the representations of the support set and query set samples, respectively. $d$ is the function to calculate the similarity score $r_{sq}$ between $\Phi_s$ and $\Phi_q$. We finally apply the cross-entropy loss to support and query pairs as follows.
\begin{equation}
\label{clsloss}
\mathcal{L}_{c l s}=\sum_{q \in Q}-\log \left(\frac{\exp \left(r_{s q}\right)}{\sum_{s' \in S} \exp \left(r_{s' q}\right)}\right).
\end{equation}

\subsection{Hierarchical compositional representations}

\begin{table}[!b]
    \vspace{-2pt}
    \caption{The Feature Encoder architecture. The initial R(2+1)D network \citep{tran2018closer} (left) and our improved version (right)}
    \label{tab:3_2_1}
    \centering
    \begin{tabular}{c|c||c|c}
            \toprule
            \multicolumn{2}{c|}{R(2+1)D} & \multicolumn{2}{c}{Ours} \\ \hline
            Layer         & Output size         & Layer         & Output size \\ \hline\hline
            $stem_1$         & ${L} \times 56 \times 56$           & $stem_1$      & ${L} \times 56 \times 56$        \\ \hline
                          &                                     & $maxpool$    & ${L} \times \textbf{28} \times \textbf{28}$        \\ \hline
            $res_2$      & ${L} \times 56 \times 56$           & $res_2$   & ${L} \times 28 \times 28$        \\ \hline
                          &                                     & \textbf{$PAM$}        & ${L} \times 28 \times 28$        \\ \hline
            $res_3$      & $\frac{L}{2} \times 28 \times 28$   & $res_3$   & $\textbf{\emph{L}} \times 14 \times 14$        \\ \hline
            $res_4$      & $\frac{L}{4} \times 14 \times 14$   & $res_4$   & $\frac{L}{2} \times 7 \times 7$  \\ \hline
            $res_5$      & $\frac{L}{8} \times 7 \times 7$     & $res_5$   & $\frac{L}{4} \times 4 \times 4$  \\ % \hline
            % $avgpool$       & $1 \times 1 \times 1$               & $avgpool$    & $1 \times 1 \times 1$  \\
            \bottomrule
        \end{tabular}
\end{table}

\begin{figure}[!t]
    \centering
    \includegraphics[width=0.28\textwidth]{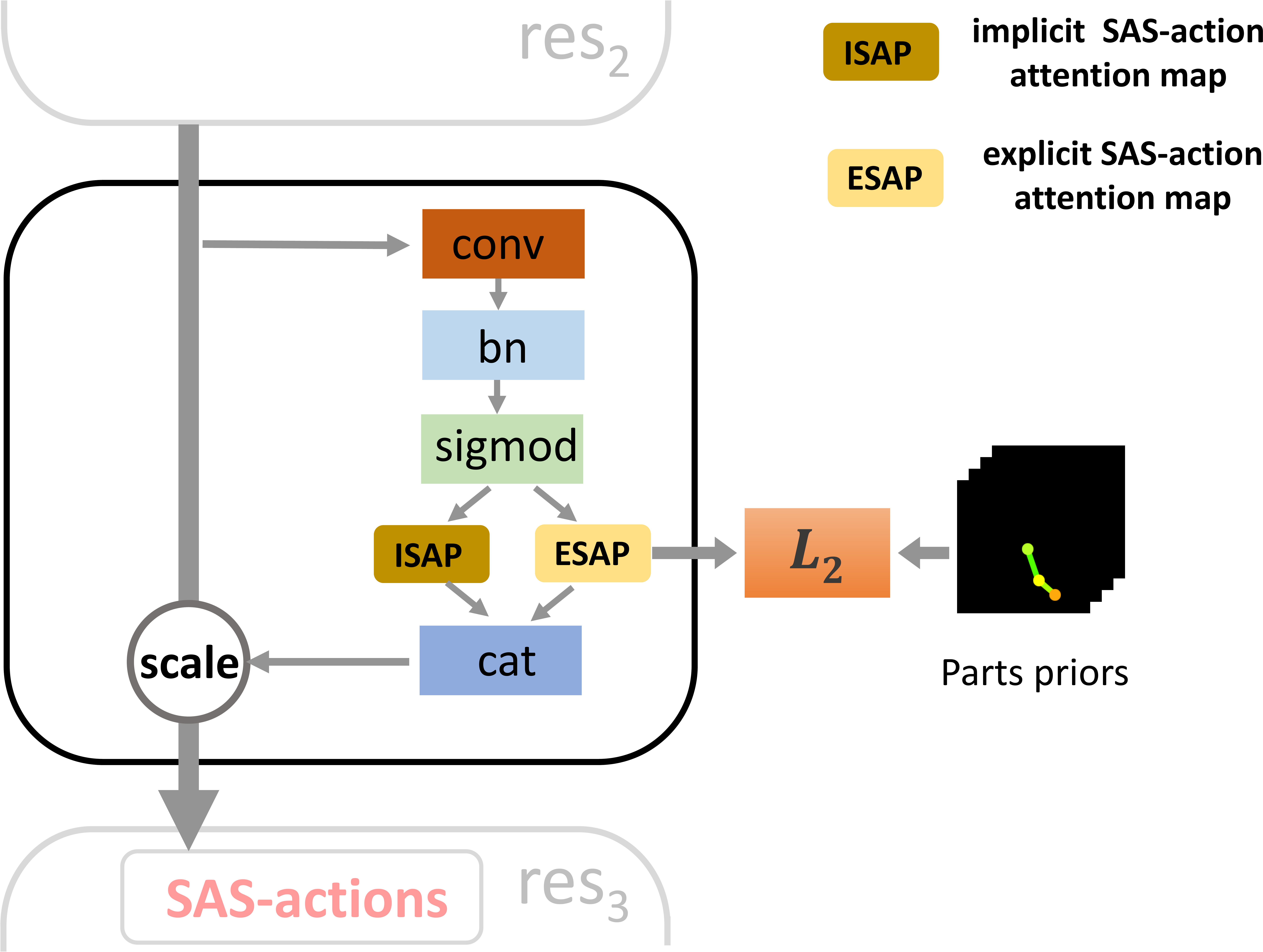}
    \caption{The Parts Attention Module (PAM) architecture. 
    The SAS-actions pay attention to various regions of interest by employing PAM,
    and especially, explicit SAS-actions focus on pre-defined body parts by parts prior
constraint.}
    \label{3_2_2}
    \vspace{-3pt}
\end{figure}

Few-shot learning aims at solving transferring knowledge from base classes to novel ones with limited training samples. However, there exist huge differences between base and novel class actions, and even no common patterns are shared for transferring between them, especially when the data of base classes are not large enough. 
Following the thought of simplifying things in cognitive science \citep{hoffman1984parts}, we divide a complicated action into several sub-actions along the temporal dimension. Besides, inspired by early works for recognizing the objects \citep{felzenszwalb2005pictorial}, we further decompose sub-actions into fine-grained SAS-actions along the spatial dimension.
After decomposing the actions in this two-level hierarchical architecture, more common patterns can be shared between the base and novel class actions, which is beneficial for transferring knowledge from base classes to novel ones. 
In our implementation, as shown in Tab. \ref{tab:3_2_1}, we adopt the efficient R(2+1)D network \citep{tran2018closer} as Feature Encoder to model sub-action representations, and meanwhile, we make some modifications for constructing hierarchical compositional representations in the following:

\begin{enumerate}
    \item Add the Parts Attention Module (PAM) at $res_2$ to assist learning SAS-actions as shown in Fig. \ref{3_2_2}.
    \item Add the spatial downsampling (MaxPool) at $stem_1$ with convolutional striding of $1 \times 2 \times 2$
    \item Remove the temporal downsampling at $res_3$
\end{enumerate}

\textbf{SAS-action representations}. The SAS-actions are decomposed from sub-actions along the spatial dimension, which pays attention to various action-relevant region information. Recent works \citep{bau2017network} about the neural network's interpretability show that each channel in specific layers can capture certain patterns of local regions. Inspired by this, we regard each channel in the specific layer as a SAS-action. Specifically, we introduce the Parts Attention Module (PAM) in Fig. \ref{3_2_2} to restrict SAS-actions to pays attention to various regions of interest (ROI). These SAS-actions consist of explicit SAS-actions and implicit SAS-actions. The ROIs of explicit SAS-actions correspond to pre-defined body parts, which are learned by human parts prior constraint, \eg, arm and leg. And the ROIs of implicit SAS-actions resort to capturing other action-related cues without any constraint. 

\begin{table*}[!h]
    \vspace{-2pt}
    \caption{The keypoints in Openpose \citep{cao2017realtime}}
    \label{keypoints}
    \centering
    \begin{tabular}{c|c||c|c||c|c||c|c||c|c||c|c}
        \toprule
        No. & keypoint & No. & keypoint & No. & keypoint & No. & keypoint & No. & keypoint & No. & keypoint \\ \hline\hline
        0 & nose & 1 & neck & 2 & right shoulder & 3 & right elbow & 4 & right wrist & 5 & left shoulder   \\ \hline
        6 & left elbow & 7 & left wrist & 8 & right hip & 9 & right knee & 10 & right ankle & 11 & left hip   \\ \hline
        12 & left knee & 13 & left ankle & 14 & right eye & 15 & left eye & 16 & right ear & 17 & left ear \\
        \bottomrule
    \end{tabular}
    \vspace{-5pt}
\end{table*}
\begin{table}[!h]
    \caption{The body parts definition}
    % \large
    \label{body parts}
    \centering
    \begin{tabular}{c|c}
        \toprule
        human parts  &  the corresponding keypoints            \\ \hline\hline
        head & [0, 14, 15, 16, 17]               \\ \hline
        left arm & [5, 6, 7]                       \\ \hline
        right arm  & [2, 3, 4]                       \\ \hline
        right leg & [8, 9, 10]                      \\ \hline
        left leg & [11, 12, 13]                    \\ \hline
        trunk & [1, 2, 5, 8, 11]                  \\ \hline
        trunk + head + left arm & \makecell[c]{[0, 1, 2, 5, 6, 7,\\ 8, 11, 14, 15, 16, 17]}            \\ \hline
        trunk + head + right arm & \makecell[c]{[0, 1, 2, 3, 4, 5,\\ 8, 11, 14, 15, 16, 17]}            \\ \hline
        trunk + head & [1, 2, 5, 8, 11, 14, 15, 16, 17]                  \\ \hline
        upper body & \makecell[c]{[0, 1, 2, 3, 4, 5,\\ 6, 7, 14, 15, 16, 17]}             \\ \hline
        lower body & [8, 9, 10, 11, 12, 13]                         \\ \hline
        left body & [0, 1, 5, 6, 7, 11, 12, 13, 15, 17]                \\ \hline
        right body & [0, 1, 2, 3, 4, 8, 9, 10, 14, 16]                  \\ \hline
        body & \makecell[c]{[0, 1, 2, 3, 4, 5, 6, 7, 8,\\ 9, 10, 11, 12, 13, 14, 15, 16, 17]} \\ \bottomrule
    \end{tabular}
    \vspace{-7pt}
\end{table}

In our implementation, we leverage Openpose \citep{cao2017realtime} to compute human pose heatmaps as a prior.
This heatmaps contain the positional information of 18 keypoints, which are listed in Table \ref{keypoints}. 
To obtain the different human body parts, we apply a per-pixel cross-channel max operation on multiple keypoint heatmaps. A detailed description of the body parts and their constituent keypoints is presented in Table \ref{body parts}.
For instance, the right arm part is defined as the set of keypoints corresponding to the right shoulder, right elbow, and right wrist.

We explore possible positions of adding PAM in the Section \ref{abst}, and we finally consider adding the PAM to right before the last layer of $res_2$ by empirical experiments. 
Therefore, we view each channel output of PAM at $res_2$ as a SAS-action representation for constructing hierarchical compositional representations.
There are 64 SAS-action representations in total, including 14 explicit SAS-actions and 50 implicit SAS-actions. We also compute 14 human parts priors $p$ to represent the corresponding body parts.
Finally, we use the Mean Square Error (MSE) loss between explicit SAS-actions $e$ and human parts priors $p$ to restrict these explicit SAS-action representations as follows. Note that the constraint is ONLY adopted in the training phase. During testing, PAM directly inference the corresponding values as predicted human parts.
\begin{equation}
\label{partloss}
\mathcal{L}_{\text {parts}}=\sum\left(e_{j}-p_{j}\right)^{2}, j=1, \ldots, 14
\end{equation}

\textbf{Sub-action representations}. The sub-actions are divided from the raw video by clustering along the temporal dimension. One simple way to achieve sub-actions is equally splitting a video sequence. However, it may cut off similar semantic information among sub-actions. To alleviate this problem, we propose an unsupervised hierarchical clustering method to obtain sub-actions with varied temporal lengths, which enforces video frames in the same sub-action to be similar while frames in different sub-actions to be different as much as possible. 

In cluster analysis \citep{rokach2005clustering}, one expects to group a set of objects so that objects in the same cluster are more similar to each other than those in other clusters. \eg, hierarchical clustering, one of the representative clustering methods, seeks to build a hierarchy of clusters. Different from conventional hierarchical clustering, the video frames clustering in our task impose the constraint of sequence continuity on input video frames. To satisfy this restriction, we adjust the nearest-neighbor chain algorithm \citep{benzecri1982construction} in agglomerative hierarchical clustering \citep{murtagh1983survey} and employ a plain algorithm to merge the closest two clusters by traversing through all neighboring clusters. More details are described in Alg. \ref{alg.1}. 
\begin{algorithm}[!b]
    \renewcommand{\algorithmicrequire}{\textbf{Input:}}
	\renewcommand{\algorithmicensure}{\textbf{Output:}}
    \caption{A plain algorithm for video frames clustering}
    \label{alg.1}
    \begin{algorithmic}[1]
        \REQUIRE video $S \in R^{T \times 3 \times H \times W}$, distance matrix $D$, $K$
        \STATE initialize partition $\left\{ {{S_1},{S_2},...,{S_T}} \right\}$, t = T
        \WHILE{$t > K$}
            \STATE $i_{min}$ = -1, $dist_{min}$ = INF
            \FOR{$i \leftarrow 1$ to $t\!\!-\!\!1$}
                \IF{$D_{i(i+1)}$ $<$ $dist_{min}$}
                \STATE $i_{min}$ = i, $dist_{min}$ = $D_{i(i+1)}$
                \ENDIF
            \ENDFOR
            \STATE \# Merge cluster $i_{min}$ and $i_{min}$\ +\ 1.
            \STATE \# Update the distance matrix $D$.
            \STATE t = t - 1
        \ENDWHILE
        \ENSURE $K$ sub-actions $\left\{ {{S_1},{S_2},...,{S_K}} \right\}$
    \end{algorithmic}
\end{algorithm}

Here, $S$ is the raw video, $D_{ij}$ is the Euclidean distance between cluster $i$ and $j$, $K$ is the sub-actions number. The total time to solve the traditional clustering by the plain algorithm is $O(n^3)$. However, the computation cost of video frames clustering is only $O(n^2)$ as we just need to repeatedly merging the two neighbor clusters for the constraint of sequence continuity. After clustering on video frames, we obtain the final sub-action representations by a series of convolution and non-linear combination on corresponding SAS-action representations.

\subsection{Distance metric}
The distance metric measures the distance between support and query samples in an embedding space. However, directly measuring global representations may destroy the temporal information and ignore local discriminative information. Besides, strictly aligning local representation sequences between videos cannot well handle time-independent cases. Here, we adopt the Earth Mover’s Distance (EMD) in the transportation problem as the distance function to match two sequences of sub-actions. In other words, we compute the minimum matching flows between two video sequences as distance function. Although EMD ignores the sequence relationship between sub-actions, the sequence information inside sub-actions is still reserved. Therefore, EMD is a good trade-off between ignoring the temporal information \citep{zhang2020few} and strictly aligning temporal information \citep{cao2020few}, which is favorable for few-shot action recognition.

\subsubsection{Earth Mover’s Distance}
The Earth Mover's Distance evaluates the dissimilarity between two multi-dimensional distributions in embedding space. In the transportation problem, suppose that $m$ suppliers $\mathbb{S}=\{p_1, p_2,...,p_m\}$ with a given amount of goods $s_i$ are required to supply $n$ consumers $\mathbb{D}=\{q_1, q_2,...,q_n\}$ with a given limited capacity $d_j$. $s_i$ is the good's amount of suppliers $i$ and $d_j$ is the capacity of consumer $j$. For each supplier-consumer pair, the cost of transporting a unit of goods from suppliers $i$ to consumer $j$ is $c_{ij}$, and the corresponding transported amount is $x_{ij}$. 
\begin{equation}
    \small
    \begin{aligned}
    x_{ij} \geq 0, \quad 1 \leq i \leq m, 1 \leq j \leq n \\ 
    \sum_{j=1}^{n} x_{i j} \leq s_i, \quad 1 \leq i \leq m \\
    \sum_{i=1}^{m} x_{i j} \leq d_j, \quad 1 \leq j \leq n \\
    \underset{x_{i j}}{\operatorname{minimize}} \sum_{i=1}^{m} \sum_{j=1}^{n} c_{i j} x_{i j}
    \end{aligned}
\end{equation}
The transportation problem is to find a minimum cost flow $x$, and meanwhile goods from the suppliers to the consumers can satisfy the consumers' demand. The minimum cost flow $x$ can be computed by solving the linear programming problem. 

In a way, transporting goods from the suppliers to the consumers is similar to matching two sequence representations, and how to define the node (supplier $i$ or consumer $j$), the cost ($c_{ij}$) and the weight ($s_j$ or $d_j$) in the few-shot action recognition will be described below.

\subsubsection{EMD for few-shot action recognition}
Here, we attempt to measure the Earth Mover's Distance between support and query samples in terms of sub-action representation sequences. Firstly, we compute the sub-action features $\Phi$ by  hierarchical compositional representations above. Then, we regard each sub-action feature as a node (like a supplier or consumer). Besides, the cost of transporting a unit of goods $c_{ij}$ is the pairwise distance between node $u_i$ and $v_j$:
\begin{equation}
c_{i j}=1-\frac{u_i^T v_j}{\left\|u_i\right\|\left\|v_j\right\|}.
\end{equation}
The weight of each node $s_i$ controls the total flow which transports goods from supplier $i$ to all consumers. $s_i=\sum_{j=1}^{n} x_{i j}$. Therefore, the node with the larger weight value indicates it is more important. Recent work \citep{zhang2020deepemd} in image tasks assigns larger weight $s_i$ to the co-occurrent nodes in two images. Following this clue, we also employ dot product between a node representation $u_i$ and the average node representation in another video $\frac{\sum_{j=1}^{K} v_j}{K}$ as the weight value $s_i$ of the node $i$.

\begin{equation}
s_i=u_i^T \cdot \frac{\sum_{j=1}^{K} v_j}{K}
\end{equation}
Therefore, the distance between two action videos $u$ and $v$ is the optimal matching cost of two corresponding sub-action representation sequences. And the similarity $r_{uv}$ between two action videos with $k$ sub-actions can be computed as follows:
\begin{equation}
r_{uv}=\sum_{i=1}^{K} \sum_{j=1}^{K}\left(1-c_{i j}\right) {x}_{i j}
\end{equation}
In our implementation, the optimal matching $x$ is differentiable by applying implicit function theorem \citep{barratt2018differentiability} on the optimality (KKT) conditions. More implementation details can be referred to the \citep{zhang2020deepemd}. Finally, we map the similarity scores to the probability distribution over various classes by employing the softmax function. The final loss contains two parts, and the hyper-parameter $\lambda$ is to balance the classification loss in Equ. \ref{clsloss} and parts loss in Equ. \ref{partloss}.
\begin{equation}
    \mathcal{L} = \mathcal{L}_{cls} + \lambda \mathcal{L}_{parts}.
    \label{lambda}
\end{equation}

\subsection{Discussions}
\textbf{Differences from DeepEMD \citep{zhang2020deepemd}.} 
Both DeepEMD and our method employ Earth Mover’s Distance (EMD) in few-shot learning, but they differ in two aspects: 1) In contrast to DeepEMD, our method learns the hierarchical compositional representations in terms of fine-grained sub-actions and SAS-actions. When training with only few labeled samples, exploring local fine-grained patterns can help provide transferable information across categories for classification. 2) DeepEMD solves the image classification task by EMD. Specifically, DeepEMD splits the building blocks with the same size along the spatial dimension and regards a local block feature as a node in transportation problems to compare counterparts in two images. Differently, we early employ EMD for action recognition in few-shot learning. And we split the sub-actions with the varying lengths along the temporal dimension and regard a clip as a node to compare fine-grained sub-actions in two videos.
In short, hierarchical compositional representations consider fine-grained movements/cues to represent videos, and employing EMD to match sub-action representations makes better use of fine-grained information. Therefore, EMD together with HCR are naturally suitable for few-shot action recognition.

\textbf{Differences from CMOT \citep{lu2021few}.} 
Both CMOT and our method measure the distance between two distributions of videos using optimal transport, but the accomplishing means are entirely different.
CMOT employs the Sinkhorn Distance to measure the distance, which is calculated by finding a regularized sparse correspondence matrix that minimizes the cost between the distributions, while our method uses Earth Mover's Distance (EMD) to measure the distance, which often involves moving mass from one distribution to another. The Sinkhorn distance can be seen as a form of regularization of EMD, but EMD provides an exact and precise measurement. Furthermore, we consider hierarchical compositional representations to better utilize fine-grained information to handle data scarcity.

\textbf{Differences from Sampler \citep{liu2022task}.} 
Sampler introduces a task-adaptive spatial-temporal video sampler for few-shot action recognition, which combines a temporal selector and a spatial amplifier to enhance video frame selection and emphasize discriminative features. Both Sampler and our method focus on spatial Regions of Interest (ROI). However, Sampler amplifies discriminative sub-regions under the guidance of saliency maps, while our method prioritizes body parts regions, which is greatly beneficial for action recognition.

\subsection{Implementation details}
Following the state-of-the-art work \citep{zhang2020deepemd} in few-shot image recognition, we adopt a two-stage training strategy: pre-training stage and fine-tuning stage. In the pre-training stage, all training split samples are trained together for learning a robust feature encoder. In the fine-tuning stage, for an n-way, k-shot problem, we randomly sample n classes in each episode and each class has k samples as the support set.
During meta-training, we sample 1 video from each class as the query set, and during meta-testing, we select 15 unlabeled videos from each class to form the query set. All models are trained on the training split. The validation split is only used for cross-validation to choose the optimal model. We calculate the average accuracy by randomly sampling 1024 episodes from test splits as the final classification accuracy. Following \citep{zhang2020few}, we report the final classification accuracy with 95 \% confidence intervals on HMDB51 and UCF101 datasets. Meanwhile, following \citep{zhu2018compound}, we directly report the final classification accuracy on Kinetics datasets.

We optimize our model by using mini-batch stochastic gradient descent (SGD) with a momentum of 0.9 and weight decay of 0.0005. Moreover, we first resize the input frames to $128 \times 128$, and then employ data augmentations \citep{wang2016temporal} including random cropping and horizontal flipping, and finally resize the cropped regions to $112 \times 112$. In the pre-training stage, the initial learning rate is 0.1 and decays every 40 epochs by 0.1; In the fine-tuning stage, the initial learning rate is 0.0001 and decays every 500 episodes by 0.5. The hyper-parameter $\lambda$ in Eq. \ref{lambda} is 1.0. Considering that sub-actions of the various lengths prevent parallel computing, we uniformly sample 4 frames in all sub-actions. 
Finally, our model is implemented by PyTorch and trained on Tesla V100 GPUs. All ablation studies results keep the equally dividing sub-actions setting for simplicity.

\section{Experiments}
In this section, we conduct comprehensive experiments to evaluate our method for few-shot action recognition. We first provide a brief introduction to the datasets. Then we make comparisons with the current state-of-the-art methods and conduct extensive ablative experiments. Finally, we visualize our model to further demonstrate the advantages of our method.

\begin{table*}[!ht]
    \caption{Comparisons with state-of-the-art methods in the 5-way setting on HMDB51 and UCF101.} 
    \label{sota}
    \centering
    \begin{tabular}{lcc|cccc}
        \toprule
        \multirow{2}{*}{Method} & \multirow{2}{*}{Pre-training} & \multirow{2}{*}{Backbone} & \multicolumn{2}{c}{HMDB51} & \multicolumn{2}{c}{UCF101} \\ \cline{4-7} 
                  &                &                    & 1-shot   & 5-shot  & 1-shot   & 5-shot  \\ \hline\hline
        FAN \citep{tan2019learning}         & ImageNet     & DenseNet121        & 50.2                  & 67.6                  & 71.8                  & 86.5            \\
        % RVN \citep{cao2021few}               & CVIU2021  & Kinetics          & 63.4 $\pm$ 0.3      & \textbf{79.7} $\pm$ \textbf{0.2}      & 88.7 $\pm$ 0.2      & \textbf{96.8} $\pm$ \textbf{0.1}  \\
        TRX \citep{perrett2021temporal}     & ImageNet     & ResNet-50                                     & -                   & 75.6                & -                   & 96.1            \\
        L2A \citep{gowda2022learn2augment}                       & ImageNet     & ResNet-50                                    & 51.9                   & \textbf{77.0}                & 79.2                   & \textbf{96.3} \\
        MTFAN \citep{wu2022motion}                     & ImageNet     & ResNet-50          & 59.0                   & 74.6                & \textbf{84.8}                   & 95.1 \\
        HCL \citep{zheng2022few}                       & ImageNet     & ResNet-50          & 59.1                   & 76.3                & 82.6                   & 94.5 \\
        HyRSM \citep{wang2022hybrid}                     & ImageNet     & ResNet-50          & 60.3                   & 76.0                & 83.9                   & 94.7 \\
        TA$^2$N \citep{li2022ta2n}                   & ImageNet     & ResNet-50          & 59.7                   & 73.9                & 81.9                   & 95.1 \\
        HCR                                 & ImageNet     & ResNet-50          & \textbf{62.5}    & 75.6   &  82.4     & 93.2  \\ 
        \hline

        GenApp \citep{mishra2018generative}  & Sports-1M     & C3D              & -                     & 52.6                  & -                     & 78.7             \\
        ProtoGAN \citep{kumar2019protogan}   & Sports-1M     & C3D              & 34.7                  & 54.0                  & 57.8                  & 80.2             \\
        
        ARN \citep{zhang2020few}             & No            & 3D-464-Conv      & 45.5                  & 60.6                  & 66.3                  & 83.1            \\
        MlSo \citep{zhang2022multi}          & No            & 3D-464-Conv      & 46.7                  & 60.3                  & 68.2                  & 87.1          \\
        HCR                                  & No            & R(2+1)D          & 48.6                  & 63.8                  & 71.8                  & 87.3          \\  
        HCR                                  & IG-65M        & R(2+1)D       & \textbf{67.5}         & \textbf{79.3}         &  \textbf{88.9}        & \textbf{95.7} \\ 
        \bottomrule
    \end{tabular}
\end{table*}

\begin{table}[!t]
    \caption{Comparisons with state-of-the-art methods in the 5-way setting on Kinetics. $\dagger$:Results from \citep{zhu2018compound}.} 
    \label{sota_k100}
    \centering
    \small
    \begin{tabular}{l|cc}
        \toprule
        \multirow{2}{*}{Method}  & \multicolumn{2}{c}{Kinetics} \\ \cline{2-3} 
                      & 1-shot(\%)   & 5-shot(\%)  \\ \hline\hline
        MatchingNet$\dagger$ \citep{vinyals2016matching}   &  53.3                 & 74.6                      \\
        MAML$\dagger$ \citep{finn2017model}                &  54.2                 & 75.3                      \\
        CMN \citep{zhu2018compound}             &  60.5                 & 78.9                      \\
        TARN \citep{bishay2019tarn}             &  66.6                 & 80.7                      \\
        ARN \citep{zhang2020few}                &  63.7                 & 82.4                      \\
        % OTAM \citep{cao2020few}                 &  73.0                 & 85.8                      \\
        TAEN \citep{ben2021taen}                &  67.3                 & 83.1                      \\
        TRX \citep{perrett2021temporal}         &  63.6                 & 85.9                      \\  
        TA$^2$N \citep{li2022ta2n}              &  72.8                 & 85.8                      \\  
        STRM \citep{thatipelli2022spatio}     &  62.9                    & 86.7                      \\  
        HyRSM \citep{wang2022hybrid}            &  73.7                    & 86.1                      \\  
        MTFAN \citep{wu2022motion}            &  74.6                 & \textbf{87.4}                      \\  
        \hline
        HCR \textit{ResNet-50}                                     &  {70.2}       & {81.4}                        \\ 
        % HCR (R(2+1)D from scratch)                                                &  53.5         & 68.9                          \\ 
        HCR \textit{R(2+1)D}                                   &  \textbf{75.7}     & 86.4                        \\ 
        \bottomrule
    \end{tabular}
\end{table}

\subsection{Datasets}
\textbf{HMDB51} \citep{kuehne2011hmdb} consists of around 7000 video clips and 51 action classes from movies, YouTube, and Web. We follow the splits protocol by \citep{zhang2020few}, where 31, 10 and 10 non-overlapping classes are selected for training, validation and testing, respectively. 
\textbf{UCF101} \citep{soomro2012ucf101} consists of over 13k clips and 27 hours of video data from 101 action classes, which contains realistic user-uploaded videos from YouTube. We follow the splits protocol by \citep{zhang2020few}, where 70, 10 and 21 non-overlapping classes are selected for training, validation and testing, respectively. 
\textbf{Kinetics} \citep{kay2017kinetics} consists of $\sim$240k training videos and 20k validation videos in 400 human action categories from YouTube. We follow the splits protocol by \citep{zhu2018compound}, where 64, 12 and 24 non-overlapping classes are selected for training, validation and testing, respectively. 

We employ three popular few-shot action recognition benchmarks to evaluate our method. These three datasets all involve daily activities containing camera motion, viewpoint change, and cluttered background, making them challenging for few-shot action recognition tasks.

\begin{table*}[!h]
    \vspace{10pt}
    \caption{Comparisons of adding PAM in different positions in the 5-way setting on HMDB51 and UCF101. (95\% confidence intervals)}
    \label{pam position}
    \centering
    \begin{tabular}{c|cccc}
        \toprule
        \multirow{2}{*}{Position} & \multicolumn{2}{c}{HMDB51} & \multicolumn{2}{c}{UCF101} \\ \cline{2-5} 
                                & 1-shot(\%)   & 5-shot(\%)  & 1-shot(\%)   & 5-shot(\%)  \\ \hline\hline
        $No$                      & 40.14 $\pm$ 0.55    & 53.51 $\pm$ 0.51    & 68.90 $\pm$ 0.62    & 85.51 $\pm$ 0.41    \\
        $stem_1$                    & 41.81 $\pm$ 0.54    & 57.49 $\pm$ 0.49    & 68.28 $\pm$ 0.64    & 84.77 $\pm$ 0.43    \\
        $res_2$                 & \textbf{45.11} $\pm$ \textbf{0.57}    & \textbf{60.29} $\pm$ \textbf{0.51}    & \textbf{70.45} $\pm$ \textbf{0.62}    & \textbf{86.56} $\pm$ \textbf{0.40}    \\
        $res_3$                 & 43.50 $\pm$ 0.58    & 60.12 $\pm$ 0.51    & 70.27 $\pm$ 0.63    & 86.21 $\pm$ 0.39    \\ 
        \bottomrule
    \end{tabular}
\end{table*}

\begin{table*}[!h]
    \caption{Comparisons of different aggregators and metrics in the 5-way setting on HMDB51 and UCF101. (95\% confidence intervals)}
    \label{metric}
    \centering
    \begin{tabular}{ccc|cccc}
        \toprule
        \multirow{2}{*}{Embedding} & \multirow{2}{*}{Aggregator} & \multicolumn{1}{l|}{\multirow{2}{*}{Metric}} & \multicolumn{2}{c}{HMDB51} & \multicolumn{2}{c}{UCF101} \\ \cline{4-7} 
                                  &                             & \multicolumn{1}{l|}{}                        & 1-shot(\%)   & 5-shot(\%)  & 1-shot(\%)   & 5-shot(\%)  \\ \hline\hline
        % \multicolumn{3}{c|}{Baseline}                                                           & 45.11 \pm 0.57        & 60.29 \pm 0.51        & 70.45 \pm 0.62        & 86.56 \pm 0.40  \\ \hline
        \multirow{4}{*}{Global}     & \multirow{2}{*}{AvgPool}      & Euclidean         & 47.79 $\pm$ 0.60        & 63.38 $\pm$ 0.51        & 71.03 $\pm$ 0.65        & 86.61 $\pm$ 0.40            \\ 
                                    &                               & Cosine            & 47.42 $\pm$ 0.58        & 63.06 $\pm$ 0.52        & 70.85 $\pm$ 0.66        & 86.59 $\pm$ 0.39            \\ \cline{2-7} 
                                    & \multirow{2}{*}{MaxPool}      & Euclidean         & 45.35 $\pm$ 0.57        & 60.13 $\pm$ 0.48        & 70.52 $\pm$ 0.64        & 86.35 $\pm$ 0.40             \\  
                                    &                               & Cosine            & 44.81 $\pm$ 0.54        & 60.09 $\pm$ 0.49        & 70.10 $\pm$ 0.64        & 86.07 $\pm$ 0.40            \\ \hline 
        Local                       & -                             & EMD               & \textbf{48.44} $\pm$ \textbf{0.59}  & \textbf{63.76} $\pm$ \textbf{0.50}  & \textbf{71.49} $\pm$ \textbf{0.64}  & \textbf{86.78} $\pm$ \textbf{0.39}            \\ 
        \bottomrule
    \end{tabular}
\end{table*}

\begin{table*}[!h]
    \caption{Comparisons of 5-way 1-shot, 2-shot, 3-shot, 4-shot, and 5-shot results on HMDB51. (95\% confidence intervals)}
    \label{shot_hmdb}
    \centering
    \begin{tabular}{lcc|ccccc}
        \toprule
        Method                  & Pre-training   & Sub-actions  & 1-shot(\%)    & 2-shot(\%) & 3-shot(\%) & 4-shot(\%)  & 5-shot(\%)    \\ \hline\hline
        \multirow{3}{*}{HCR}    &                & 8            & 48.44 $\pm$ 0.59    & 55.62 $\pm$ 0.55    & 59.76 $\pm$ 0.53    & 62.43 $\pm$ 0.51    & 63.76 $\pm$ 0.50              \\ 
                                & \checkmark     & 8            & 66.53 $\pm$ 0.67    & 73.00 $\pm$ 0.59    & 76.34 $\pm$ 0.53    & 77.63 $\pm$ 0.48    & 78.91 $\pm$ 0.46              \\ 
                                & \checkmark     & 12           & 67.47 $\pm$ 0.68    & 73.56 $\pm$ 0.59    & 76.59 $\pm$ 0.53    & 78.10 $\pm$ 0.49    & 79.28 $\pm$ 0.46              \\ 
        \bottomrule
    \end{tabular}
\end{table*}

\begin{table*}[!h]
    \caption{Comparisons of 5-way 1-shot, 2-shot, 3-shot, 4-shot, and 5-shot results on UCF101. (95\% confidence intervals)}
    \label{shot_ucf}
    \centering
    \begin{tabular}{lcc|ccccc}
        \toprule
        Method                  & Pre-training   & Sub-actions  & 1-shot(\%)    & 2-shot(\%) & 3-shot(\%) & 4-shot(\%)  & 5-shot(\%)    \\ \hline\hline
        
        \multirow{3}{*}{HCR}    &                & 8            & 71.49 $\pm$ 0.64    & 80.05 $\pm$ 0.51    & 83.33 $\pm$ 0.46    & 85.20 $\pm$ 0.41    & 86.78 $\pm$ 0.39              \\ 
                                & \checkmark     & 8            & 87.36 $\pm$ 0.53    & 91.99 $\pm$ 0.37    & 93.39 $\pm$ 0.34    & 94.16 $\pm$ 0.30    & 94.94 $\pm$ 0.26              \\ 
                                & \checkmark     & 16           & 88.85 $\pm$ 0.52    & 92.96 $\pm$ 0.36    & 94.25 $\pm$ 0.33    & 94.99 $\pm$ 0.29    & 95.68 $\pm$ 0.25              \\
        \bottomrule
    \end{tabular}
\end{table*}

\begin{table*}[!h]
    \caption{Comparisons of 5-way 1-shot, 2-shot, 3-shot, 4-shot, and 5-shot results on Kinetics.}
    \label{shot_k100}
    \centering
    \begin{tabular}{lcc|ccccc}
        \toprule
        Method                  & Pre-training   & Sub-actions  & 1-shot(\%)    & 2-shot(\%) & 3-shot(\%) & 4-shot(\%)  & 5-shot(\%)    \\ \hline\hline
        \multirow{3}{*}{HCR}    &                & 8            & 52.71     & 60.40     & 63.93     & 66.64     & 68.29              \\ 
                                % &                & \textcolor{mygray}{24}            & \textcolor{mygray}{53.53}     & \textcolor{mygray}{61.04}     & \textcolor{mygray}{64.39}     & \textcolor{mygray}{67.23}     & \textcolor{mygray}{68.88}              \\ 
                                & \checkmark     & 8            & 73.32     & 79.64     & 82.19     & 83.98     &  84.56             \\ 
                                & \checkmark     & 24            & 75.65     & 81.76     & 84.17     & 85.72     & 86.37              \\
        \bottomrule
    \end{tabular}
\end{table*}

\subsection{Comparison with previous works}
\label{exp}
As depicted in Tables \ref{sota} and \ref{sota_k100}, we present comprehensive comparisons with existing state-of-the-art methods utilizing both 2D ConvNets and 3D ConvNets architecture on HMDB51, UCF101, and Kinetics datasets. Notably, our HCR measures similarity at the sub-action level using 3D ConvNets. Therefore, for fair comparisons with methods pre-trained on ImageNet, we adopt a 2D ConvNet (Resnet50) backbone with specific adjustments, including adjusting sub-action lengths, removing spatial downsampling, altering input resolution, incorporating 2D Parts Attention Module (PAM), and exploring new hyper-parameter spaces. Detailed descriptions of these modifications are available in the Appendix.

Employing the 2D ConvNets architecture, our HCR achieves the highest accuracy of 62.5\% on HMDB51 and competitive results of 82.4\% on UCF101 in the 5-way, 1-shot setting, outperforming existing methods such as L2A and TA$^2$N. An interesting finding is the slight accuracy improvement observed in the 5-shot setting compared to the 1-shot setting. The possible reason is that the 5-shot evaluation was directly conducted using models trained with the 1-shot setting following the setting in DeepEMD \citep{zhang2020deepemd}.

Regarding the use of 3D ConvNets, we leverage a more efficient R(2+1)D network \citep{tran2018closer} as the backbone, initially pre-trained on the unsupervised IG-65M dataset \citep{ghadiyaram2019large}. This setting proves particularly effective in the few-shot setting on both HMDB51 and UCF101, verifying the efficacy of large-scale data in tackling the data scarcity issue inherent in few-shot learning.

Compared to existing methods, our method demonstrates more significant accuracy improvement on HMDB51 than on UCF101, as shown in Table \ref{sota}. This can be attributed to the HMDB51 dataset emphasis on pose-related action recognition \citep{li2020pas}, which works well with the strengths of our hierarchical compositional representations in leveraging pose information. Conversely, the UCF101 dataset, characterized by more distant views, faces challenges in obtaining accurate keypoint information.

Following the Kinetics-CMN protocol \citep{zhu2018compound} as shown in Table \ref{sota_k100}, our HCR shows competitive results when utilizing the Imagenet pre-trained ResNet-50 architecture, comparable to TRX \citep{perrett2021temporal} and STRM \citep{thatipelli2022spatio}. Notably, using the IG-65M pre-trained R(2+1)D architecture, our HCR outperforms the current state-of-the-art, MTFAN \citep{wu2022motion}, especially under the 1-shot setting. This further establishes the superiority of our method in few-shot action recognition.

\subsection{Ablation study}
\label{abst}

\textbf{The position of adding PAM}. Table \ref{pam position} shows the effect of adding PAM in different positions. We discuss the cases that adding PAM to right before the last layer of $stem_1$, $res_2$, $res_3$ as shown in Fig. \ref{3_2_2}. Compared to the case without parts constraint, adding PAM on $res_2$ and $res_3$ show a significant performance improvement, while on $stem_1$ has no obvious improvement. One possible cause is that the receptive field is relatively small in the shallow layer, and there exist conflicts between small-scale low-level SAS-action representations and large-scale body parts priors.

\textbf{Distance metric}. Table \ref{metric} compares the EMD and traditional metrics. Here, we compare two common distance metrics in few-shot learning: 1) Euclidean distance in Prototypical Network \citep{snell2017prototypical}. 2) Cosine distance in Matching Network \citep{vinyals2016matching}. Moreover, we also compare common pooling methods to aggregate global video-level representations: average pooling (AvgPool) and max pooling (MaxPool). 
We observe that the EMD metric outperforms all traditional metrics. The EMD computes the distance of sub-action representation sequences among videos, demonstrating its superiority for few-shot action recognition.
Moreover, we discover that the AvgPool has a better performance than MaxPool. One possible reason is that although the MaxPool can achieve the most discriminative features, it may ignore less discriminative features, which are still useful for few-shot action recognition.
Furthermore, the accuracy of cosine distance is quite close to that of Euclidean distance, both of which well estimate the similarity between action videos.
% these salient features are hard to distinguish in the category space. 
% demonstrates the effectiveness of hierarchical compositional representations again.
% To verify the advantage of our method

\begin{figure*}[!h]
    \vspace{4pt}
    \centering
    \subfloat[HMDB51 dataset]{
        \includegraphics[width=0.24\linewidth]{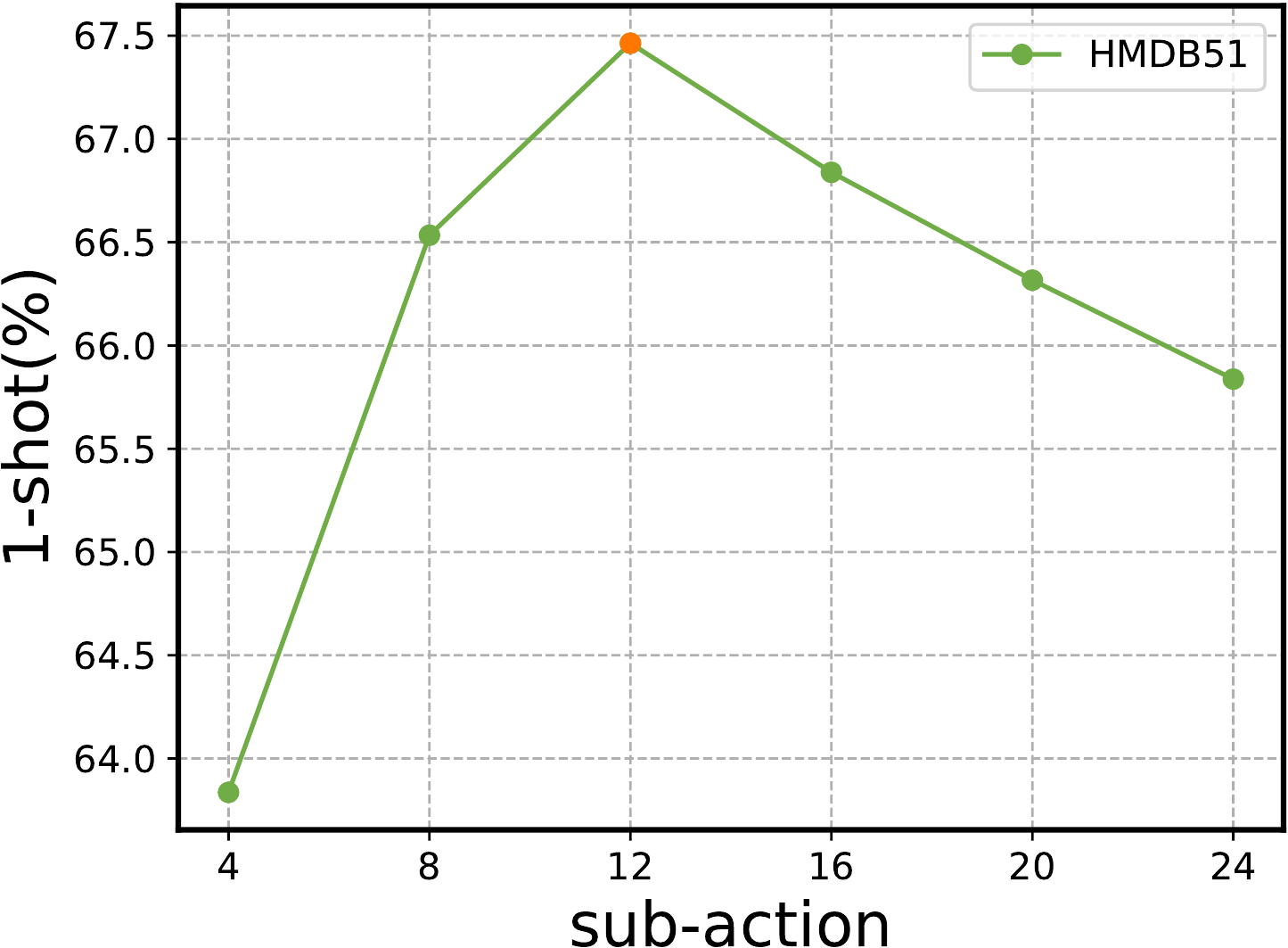}%
        \label{fig_first_case}
        }
    \hfil
    \subfloat[UCF101 dataset]{
        \includegraphics[width=0.24\linewidth]{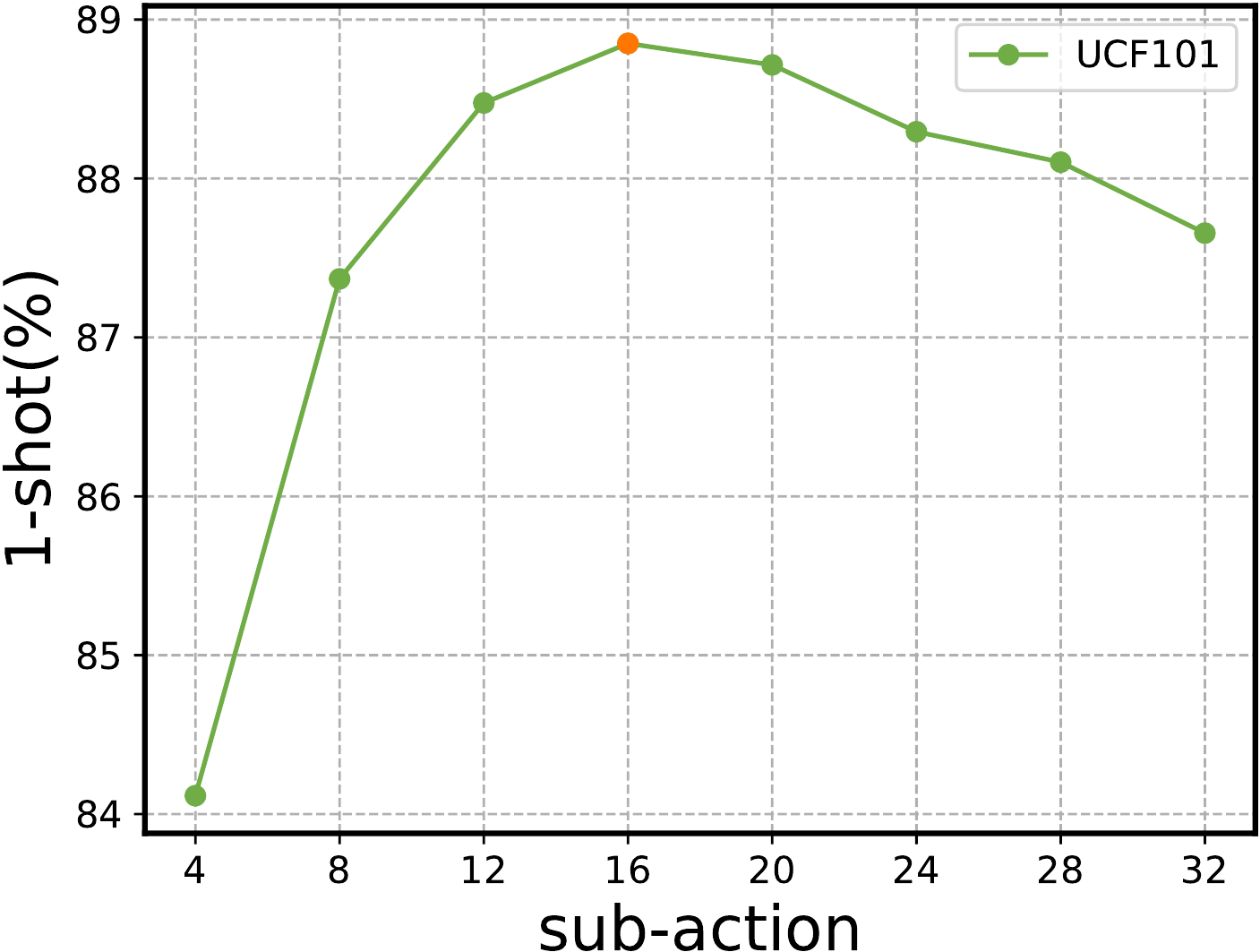}%
        \label{fig_second_case}
        }
    \hfil
    \subfloat[Kinetics dataset]{
        \includegraphics[width=0.24\linewidth]{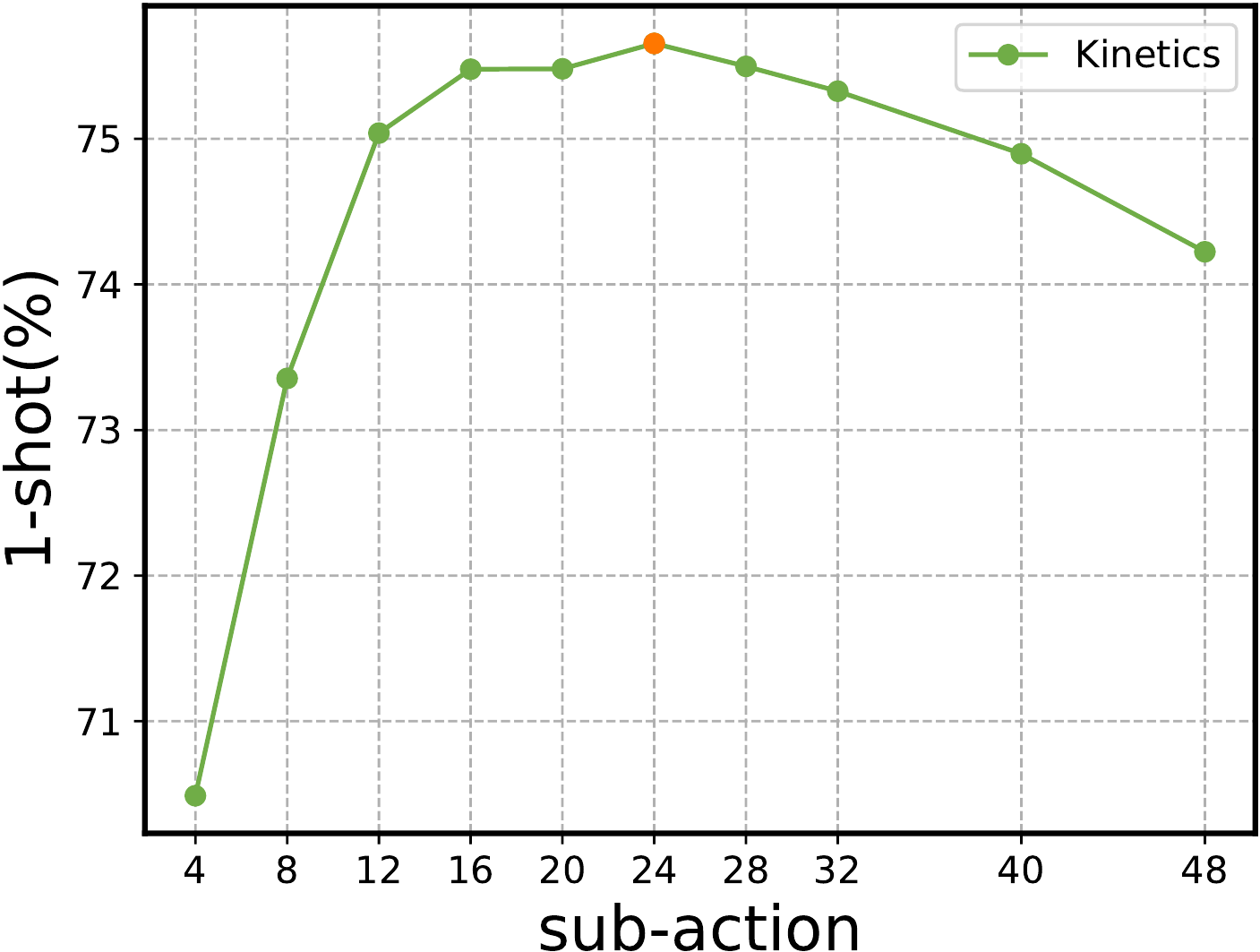}%
        \label{fig_third_case}
        }
    \caption{Accuracy comparisons of various sub-action numbers in the 5-way 1-shot setting on HMDB51 (left), UCF101 (middle) and Kinetics (right)}
    \label{subaction}
    \vspace{10pt}
\end{figure*}

\textbf{The number of sub-actions}. Figure 4 presents accuracy comparisons across various sub-action numbers. It is observed that the recognition accuracy initially improves as the number of sub-actions increases. However, a subsequent decline in accuracy is observed upon further increasing the sub-action numbers. This trend is likely due to excessive sub-actions containing redundant information, which complicates the optimization of the EMD metric through numerous similar sub-action pairs. Notably, our model demonstrates optimal recognition accuracies of 67.5\%, 88.9\%, and 75.7\% on HMDB51, UCF101, and Kinetics, respectively, with 12, 16, and 24 sub-actions. The variation in optimal sub-action numbers across these datasets can be attributed to their differing average video lengths.

\textbf{The number of shots}. The results detailed in Tables \ref{shot_hmdb}, \ref{shot_ucf}, and \ref{shot_k100} illustrate the impact of the number of shots in the 5-way setting for HMDB51, UCF101, and Kinetics, respectively. Across all models, there is an obvious improvement in accuracy with the increase in the number of shots for novel classes. Furthermore, models pre-trained on the IG-65M dataset \citep{ghadiyaram2019large} significantly outperform those trained from scratch. This underscores the value of leveraging knowledge from other domains to effectively address the challenge of data scarcity inherent in the few-shot learning.

\begin{figure}[!t]
\scriptsize
    \centering
    \includegraphics[width=0.44\textwidth]{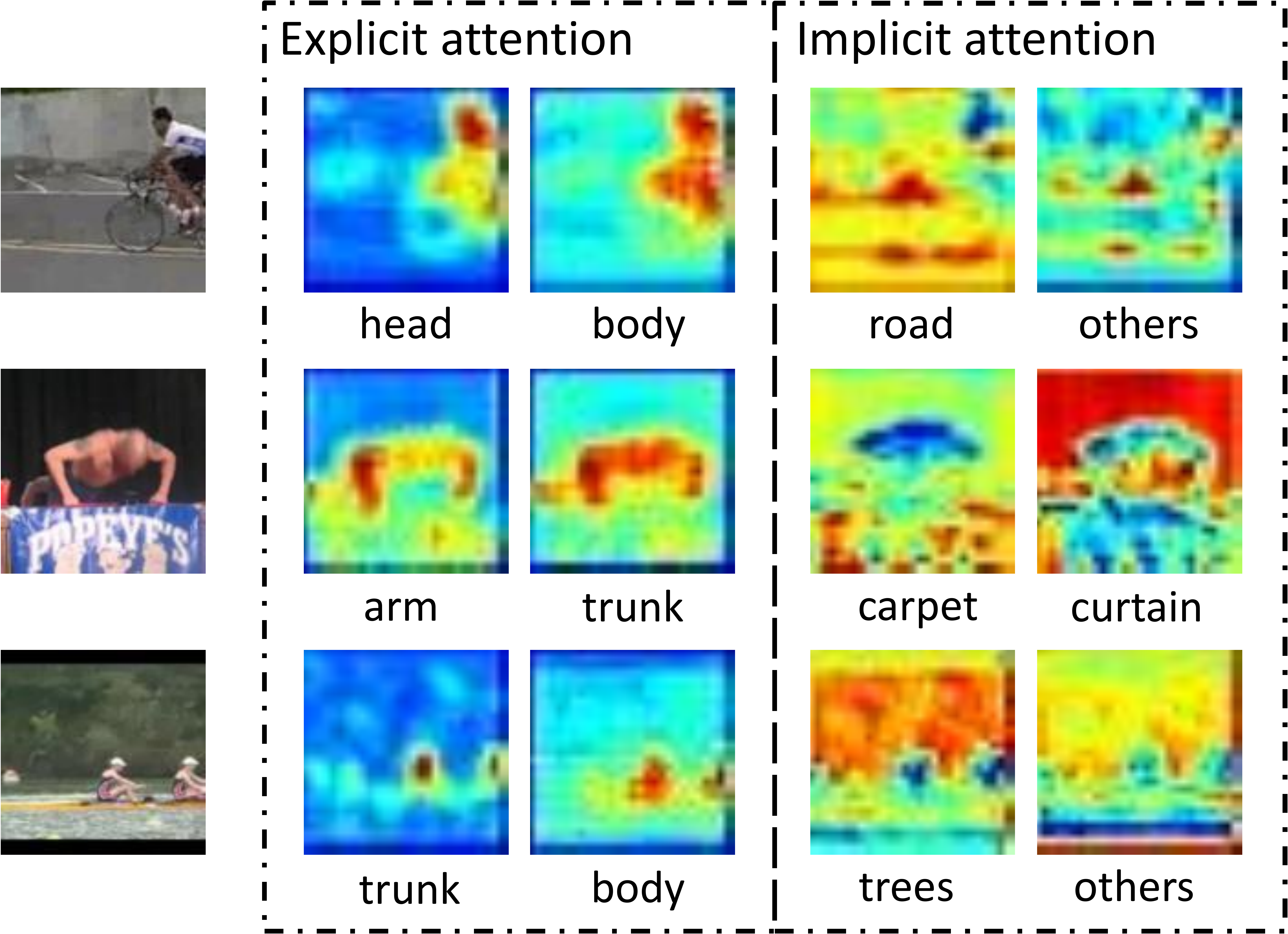}
    \caption{The visualization results. The PAM restricts
SAS-actions to pays attention to specific regions of interest.}
    \label{vis}
    \vspace{-5pt}
\end{figure}

\subsection{Visualization}
We suppose that explicit SAS-actions pay attention to pre-defined human body parts while implicit SAS-actions pay attention to other action-relevant cues like context. To verify this, Figure. \ref{vis} visualizes the predicted SAS-actions attention maps (in Fig. \ref{3_2_2}) to show what regions the SAS-actions focus on. 
For the action "ride a bike", explicit SAS-actions concentrate its attention on local body regions, \eg, head or body, while the regions implicit SAS-actions focused on are hard to express, including roads and other action-relevant contexts. For the action "push up", explicit SAS-actions pay attention to arm or trunk, while implicit SAS-actions focus on carpet or curtain. And for the action "rowing", explicit SAS-actions focus on trunk or body, while implicit SAS-actions pay attention to trees and other action-relevant contexts.

\section{Conclusion}
This paper proposes a novel approach by learning hierarchical compositional representations and employing Earth Mover’s Distance for few-shot action recognition. 
We discover that hierarchical compositional representations can well exploit potential fine-grained patterns, which helps to generalize patterns from base actions to recognize novel actions.
Further, we explore adopting Earth Mover's Distance to measure the similarity between video samples and verify its effectiveness for comparing fine-grained patterns. 
Extensive experiments show that our method achieves the state-of-the-art results for few-shot action recognition on popular datasets. 
% A future direction is to study generalized few-shot action recognition algorithms.

\section*{Acknowledgments}
This work is partially supported by National Key R\&D Program of China (No. 2021YFC3310100), National Natural Science Foundation of China (No. 62176251), and Youth Innovation Promotion Association CAS.

\bibliographystyle{model2-names}
\bibliography{egbib}

\end{document}